\documentclass[conference]{IEEEtran}
\usepackage{microtype}
\usepackage{amsmath}
\usepackage{amsfonts}
\usepackage{mathtools}
\usepackage{scrextend}
\usepackage{tabu}
\usepackage{multirow}
\usepackage{subcaption}
\usepackage{graphicx}
\usepackage{epstopdf}

\newcommand{\norm}[1]{\left\lVert#1\right\rVert}

\begin{document}
\title{Sparse Multidimensional Patient Modeling using Auxiliary Confidence Labels}

\author{\IEEEauthorblockN{Eric Heim, Milos Hauskrecht}
\IEEEauthorblockA{University of Pittsburgh, Department of Computer Science, Pittsburgh, PA, USA, Email: \{eric,milos\}@cs.pitt.edu}}

\maketitle

\begin{abstract}
\label{abstract}
In this work, we focus on the problem of learning a classification model that performs inference on patient Electronic Health Records (EHRs).  Often, a large amount of costly expert supervision is required to learn such a model.  To reduce this cost, we obtain \emph{confidence labels} that indicate how sure an expert is in the class labels she provides.  If meaningful confidence information can be incorporated into a learning method, fewer patient instances may need to be labeled to learn an accurate model.  In addition, while accuracy of predictions is important for any inference model, a model of patients must be interpretable so that clinicians can understand how the model is making decisions.  To these ends, we develop a novel metric learning method called \textbf{C}onfidence b\textbf{A}sed \textbf{ME}tric \textbf{L}earning (CAMEL) that supports inclusion of confidence labels, but also emphasizes interpretability in three ways.  First, our method induces sparsity, thus producing simple models that use only a few features from patient EHRs.  Second, CAMEL naturally produces confidence scores that can be taken into consideration when clinicians make treatment decisions.  Third, the metrics learned by CAMEL induce multidimensional spaces where each dimension represents a different ``factor'' that clinicians can use to assess patients.  In our experimental evaluation, we show on a real-world clinical data set that our CAMEL methods are able to learn models that are as or more accurate as other methods that use the same supervision.  Furthermore, we show that when CAMEL uses confidence scores it is able to learn models as or more accurate as others we tested while using only 10\% of the training instances.  Finally, we perform qualitative assessments on the metrics learned by CAMEL and show that they identify and clearly articulate important factors in how the model performs inference.
\end{abstract}

\IEEEpeerreviewmaketitle

\section{Introduction}
\label{sec:introduction}
As recent technological advancements become more integrated into the practice of clinical medicine, more opportunities arise to support clinicians when they make important decisions in patient care.  
This has lead to the advent of \emph{Clinical Decision Support Systems} (CDSSs), which are computer systems that use data to aid clinicians in making clinical decisions.  
CDSSs can simply act as a portal for clinicians to access relevant information, but can perform much more sophisticated tasks such as providing suggested treatment options or warning of dangerous drug interactions.  
For a CDSS to accomplish such inference tasks, it requires a meaningful model of how previously observed patients relate to new patients.
To build such a model, data regarding previous patients and task-specific supervision on those patients is required. 
Fortunately, \emph{Electronic Health Records} (EHRs) are being adopted by more and more health care providers \cite{jha2011progress,charles2013adoption}.  
EHRs provide data that uniquely characterizes different patients in an easily accessible form.  
For supervision, clinicians themselves can provide quality feedback if explicitly prompted for it.  
By combining these two sources of information, an insightful inference model of patients can be built using supervised learning techniques.




Much of the previous work in creating patient models from supervision leverage standard classification methods \cite{guyon2002gene, el2002support, eom2008aptacdss, statnikov2008comprehensive}.  
Here, supervision appears in the form of class labels (e.g. the patient is at risk for a condition or not), and the learned inference models output a predicted class label when given an unseen patient.  
For CDSSs, these predictions can be used to alert clinicians of important information that supports decision making.
However, there are practical issues with standard classification models for use in CDSSs.   First, it is vital that a clinician is able to understand how a CDSS comes to conclusions \cite{berner2007clinical}. Otherwise, the clinician may not trust the model due to lack of clear reasoning.  Many standard classification methods focus solely on maximizing some measure of classification accuracy without any focus on learning a model that can be easily interpreted by humans.  Consequently, clinicians may not be able to understand why they are being alerted, even the classifier is accurate. 

Another practical concern lies in the cost of obtaining sufficient clinical supervision to learn an accurate classification model.  
Because the expertise of a clinician is valuable, the cost of obtaining clinical supervision is substantially more than obtaining feedback from the layman.
Compounding this cost is that clinicians must spend a substantial amount of time to consider multiple, interacting factors before providing feedback.
If standard classification methods are to be used, clinicians would be prompted for a class label after considering a patient.  
However, class labels convey only a simple notion of how patients relate, despite the fact that clinicians have more in-depth knowledge about the patient that they could provide. 
Thus, often a large amount of labeled instances needed to learn accurate classifiers for more complex inference tasks.  All of these factors together make the cost of learning an accurate classification model from class labels alone an expensive endeavor.


In consideration of these issues, this work proposes a novel \emph{metric learning} method called \textbf{C}onfidence-b\textbf{A}sed \textbf{ME}tric \textbf{L}earning (CAMEL).  CAMEL was designed with a specific emphasis on learning human-interpretable models of patients.  Our method produces sparse models that use only the relevant patient information when modeling disease.  As a result, clinicians can easily identify which features are used when making inferences.  Also, metrics produced by CAMEL naturally induce confidence scores that indicate how confident the model is when making inferences.  Clinicians can take these scores into consideration when making important treatment decisions.  Finally, our method learns a parametric metric that projects patients into multidimensional space of disease where each dimension in the learned metric space is as a separate ``factor'' in how the model reasons about patients.  This contrasts with many standard classification methods that project data objects to a single dimension.  Clinicians can view what features influence these factors and interpret how the model is making inferences.

To reduce the cost of obtaining expert supervision, we formulate a version of CAMEL that can leverage \emph{confidence labels} that indicate how sure a labeler is in a given class label.  Our use of confidence labels is motivated by the fact that when tasked with providing supervision, clinicians spend their time mostly on considering the patient EHRs themselves.  Once they learn what is needed to produce a class label for a patient, providing a confidence label requires a relatively short amount of additional time and effort. If this confidence information provides additional, useful insight into how patients relate to classes, then our method that utilizes confidence labels, CAMEL-CL, is able to learn more accurate classification models with fewer labeled patients.  In doing so, CAMEL-CL can be used to reduce the effort required from the labeler, and in turn, decrease the cost of obtaining expert supervision.  

The remainder of this paper is organized as follows.  In Sec. \ref{sec:methodology} we review pertinent background information in metric learning and then formally define our two CAMEL methods.  In Sec. \ref{sec:relWork} we briefly review previous, related work.  In Sec. \ref{sec:experiments} we perform an experimental evaluation of CAMEL and CAMEL-CL, assessing the models they learn first quantitatively by comparing their accuracy to models learned by other methods, then qualitatively in terms of their interpretability.  Finally, in Sec. \ref{sec:conclusion} we conclude and note directions of future work.



\section{Methodology}
\label{sec:methodology}
Before defining our proposed methods, we begin this section by reviewing basic concepts in Mahalanobis distance metric learning and comparing the models they learn to those produced by traditional linear models.  With this background, we introduce our method, CAMEL, and discuss how it can utilize both class and confidence labels to learn a model of patients.

\subsection{Mahalanobis Distance Metric Learning}
A \emph{metric} is a function $d: \mathcal{X} \times \mathcal{X} \to \mathbb{R}$ that defines a measure of distance between pairs of objects.  More specifically, in order to be a metric, $d$ must satisfy four conditions (non-negativity, distinguishability, symmetry, and triangle inequality) that embody intuitive properties that traditional notions of distance have.
In \emph{metric learning} \cite{yang2006distance,kulis2012metric,bellet2013survey}, the goal is to learn a metric from data.  Most commonly, supervision is used to guide this process.  For example, supervision can indicate which objects are in some sense ``similar'' and which ones are ``different''.  With this feedback, a metric learning method could learn $d$ such that similar objects are closer to each other than different objects.  One of the most popular classes of metric learning methods are those that learn a squared \emph{Generalized Mahalanobis Distance Metric} (GMDM) of the form:
\begin{equation}
  d_\mathbf{M}^2\left(\mathbf{x}_i, \mathbf{x}_j\right) = \left(\mathbf{x}_i - \mathbf{x}_j\right)^{T}\mathbf{M}\left(\mathbf{x}_i - \mathbf{x}_j\right)
\label{eq:Mahal}
\end{equation}  
\noindent Here, $\mathcal{X} = \mathbb{R}^m$.  The metric defined in \eqref{eq:Mahal} is parameterized by $\mathbf{M} \in \mathbb{R}^{m \times m}$. If $\mathbf{M}$ is positive semidefinite (PSD) (i.e has no negative eigenvalues), then \eqref{eq:Mahal} is a \emph{pseudometric}, which satisfies all the properties of a metric except distinguishability. 

Intuition into GMDMs can be gained by factoring the square matrix parameter $\mathbf{M} = \mathbf{L}^{T}\mathbf{L}$, where $\mathbf{L} \in \mathbb{R}^{m' \times m}$. One can then rewrite \eqref{eq:Mahal} as a function of $\mathbf{L}$:
\begin{equation}
  \begin{array}{rl}
    d_\mathbf{L}^2\left(\mathbf{x}_i, \mathbf{x}_j\right) & = \left(\mathbf{x}_i - \mathbf{x}_j\right)^{T}\mathbf{L}^T\mathbf{L}\left(\mathbf{x}_i - \mathbf{x}_j\right)\\
 & = \left(\mathbf{L}\mathbf{x}_i - \mathbf{L}\mathbf{x}_j\right)^{T}\left(\mathbf{L}\mathbf{x}_i - \mathbf{L}\mathbf{x}_j\right) \\
& = \norm{\mathbf{L}\mathbf{x}_i - \mathbf{L}\mathbf{x}_j}_2^2 = d_2^2\left(\mathbf{L}\mathbf{x}_i,\mathbf{L}\mathbf{x}_j\right)
  \end{array}
\label{eq:MahalFactored}
\end{equation}  
The last line of \eqref{eq:MahalFactored} shows that a GMDM is equivalent to the standard Euclidean distance metric squared after the $m$-dimensional objects are transformed into an $m'$-dimensional metric space.  GMDM learning (henceforth, metric learning) methods use $\mathbf{M}$, or equivalently $\mathbf{L}$, as a parameter to be learned. In doing so, they create a linear transformation of objects from a given feature space to a metric space in a data-driven way.

One of the tasks metric learning methods are used for is classification.  For this, metrics are learned that can be used to determine what class an unobserved data object belongs to by comparing it to observed, labeled instances.  Many standard classification methods (SVM, logistic regression, etc.) learn a linear transformation into a single dimension.  Learning such a simple model can often have practical limitations.  First, for some domains, objects of different classes cannot be separated by a single-dimensional linear transformation, making many standard methods inappropriate.  Linear classifiers often compensate for this by including a manually chosen feature mapping (or a kernel) to first map the objects into a space that can be separated by the single-dimensional transformation.  However, which feature mapping is suited for a problem is often unclear, and choosing an appropriate mapping can be time consuming.  Even if an appropriate feature mapping is chosen, the resulting model may be difficult to interpret.  Because the features are being weighed along one dimension, it is impossible to tell if certain features are important on their own, or in tandem with others, and sometimes whether some features are important at all.  Our method, CAMEL, not only implicitly learns an appropriate feature mapping by transforming data objects into a metric space, but does so by combining features in different combinations, each of which can be interpreted independently to understand how the model is making inferences.  In the remainder of this section, we outline our method and highlight certain design decisions that enhance interpretability.

\subsection{Confidence-Based Metric Learning from Class Labels}
We begin by more formally defining our problem setting.  Let $\mathcal{D} = \{\left(\mathbf{x}_1,y_1,c_1\right),...,\left(\mathbf{x}_n,y_n,c_n\right) \in \left(\mathcal{X},\mathcal{Y},\mathcal{C}\right)^n\}$ be a set of observed data.  Let $\mathcal{X} = \{\mathbf{x}_1,...,\mathbf{x}_n\in \mathbb{R}^m\}$ be a collection of $n$ data objects represented by $m$-dimensional real vectors.  Let $\mathcal{Y} = \{y_1,...,y_n\in \{0,1\}\}$ and $\mathcal{C} = \{c_1,...,c_n\in \left[0,1\right]\}$ be binary and confidence labels, respectively, that correspond to data objects. In our problem setting, each $\mathbf{x}_i$ is a patient instance represented by features drawn from EHR data.  The corresponding $y_i$ is a class label gathered from a clinician (e.g. a positive or negative diagnosis), and $c_i$ is a confidence label indicating how confident she is in $y_i$.  In this work, we consider binary class labels and confidences in $\left[0,1\right]$, though much of the subsequent can easily be extended to other settings.

We wish to learn a metric parameter $\mathbf{L}$ from observed patient data that can be used to accurately predict the class labels of unobserved patient instances.  To this end, we begin by defining a measure of similarity between objects, given $\mathbf{L}$:
\begin{equation}
k_\mathbf{L}\left(\mathbf{x}_i, \mathbf{x}_j\right) = \mathrm{exp}\left(-d_\mathbf{L}^2\left(\mathbf{x}_i, \mathbf{x}_j\right)\right)
\label{eq:GuassianMetric}
\end{equation}
\noindent Equation \eqref{eq:GuassianMetric} is an application of the Gaussian kernel function (also known as the radial basis function) \cite{scholkopf2002learning} that measures how similar two objects are.  If the distance between $\mathbf{x}_i$ and $\mathbf{x}_j$ is zero, then \eqref{eq:GuassianMetric} assigns the pair a similarity of one.  As objects become farther apart, their similarity quickly goes to zero.  The Gaussian kernel function is a popular choice in kernelized learning algorithms because it can model a large class of functions.  However, it is normally parameterized by a bandwidth that influences how quickly similarities decay towards zero.  This parameter greatly affects the performance of the methods in which the Gaussian kernel is used, and needs to be validated for use in traditional classification methods.  In \eqref{eq:GuassianMetric}, the bandwidth parameter is absorbed into the learned parameter $\mathbf{L}$.  As a result, our method not only learns a transformation, but also the bandwidth of a Gaussian kernel. 

With this measure of similarity, we can define the relationship a patient instance has with others.  Most importantly, we can define how similar an patient is to those with class label $y$:
\begin{equation}
S_{\mathbf{L}}^y\left(\mathbf{x}_i\right) = \frac{1}{|\mathcal{X}^y_{\mathbf{x}_i}|} \sum_{\forall \mathbf{x}_j \in \mathcal{X}^y_{\mathbf{x}_i}} k_\mathbf{L}\left(\mathbf{x}_i, \mathbf{x}_j\right)
\label{eq:ClassSim}
\end{equation}
\noindent Here, $\mathcal{X}^y_{\mathbf{x}_i} = \{\mathbf{x}_j \in \mathcal{X}: y_j = y \land \mathbf{x}_j \neq \mathbf{x}_i\}$.  In essence, \eqref{eq:ClassSim} is the mean similarity $\mathbf{x}_i$ has with all observed objects with label $y$, excluding itself.  This \emph{similarity score} measures how similar a patient is to observed patient instances of a single class.  However, this score is independent of a object's relationship to other classes.  For this, we formulate a \emph{confidence score}:
\begin{equation}
C_{\mathbf{L}}^y\left(\mathbf{x}_i\right) = \frac{S_{\mathbf{L}}^y\left(\mathbf{x}_i\right)}{S_{\mathbf{L}}^y\left(\mathbf{x}_i\right) + S_{\mathbf{L}}^{\tilde{y}}\left(\mathbf{x}_i\right)}
\label{eq:ClassConf}
\end{equation}
\noindent In the binary class case, $\tilde{y}$ is zero if $y$ is one, and one if $y$ is zero (the complement of $y$).  In the multi-class case, $\tilde{y}$ is all class labels other than $y$.  Equation \eqref{eq:ClassConf} can be interpreted as a class conditional probability that an object is a member of a class given a metric parameter $\mathbf{L}$.  As such, confidence scores define both a criteria for learning an $\mathbf{L}$ that fits to observed data, and a way to predict class labels on unobserved patient instances once $\mathbf{L}$ is learned.  Like other classifiers that define conditional probability functions, inference can be done by putting a threshold on the confidence score of an unobserved instance (e.g. if $C_{\mathbf{L}}^1\left(\mathbf{x}_i\right) > 0.4$, then the predicted label $\hat{y}_i$ is one, otherwise, it is zero).  In addition, area under the Receiver Operating Characteristic curve (AUROC) can be found using confidence scores on patient instances.

In order to find a metric that models class membership of observed patient instances well, one could maximize $C_{\mathbf{L}}^{y_i}\left(\mathbf{x}_i\right)$ directly for all observed patients.  However, doing so leads to a difficult non-convex optimization problem.  To avoid this  source of nonconvexity, we maximize an approximation of the confidence scores for the observed data:
\begin{equation}
    \displaystyle \max_{\mathbf{L}} \sum_{i=1}^n S_{\mathbf{L}}^{y_i}\left(\mathbf{x}_i\right) - S_{\mathbf{L}}^{\tilde{y}_i}\left(\mathbf{x}_i\right) 
    \label{eq:CAMELnoReg}
\end{equation}
\noindent  In \eqref{eq:CAMELnoReg} we maximize the similarity each observed data object has with its observed class label, while minimizing the the similarity it has with the opposite class.  The objective ``pulls'' all observed objects of the same class towards each other, and ``pushes'' all objects of different classes away.  In doing so, \eqref{eq:CAMELnoReg} increases the numerator of \eqref{eq:ClassConf}, while decreasing a term the denominator, and by doing so, approximates learning a metric where the confidence scores of the observed data is high. 

Unfortunately, \eqref{eq:CAMELnoReg} can result in solutions where observed objects of the same class are projected to nearly the same point in the metric space, while observed objects of different classes are infinitely far apart.  This leads to models that can drastically over-fit to the observed data.  To combat this, we include $l$-1 norm regularization into the objective:
\begin{equation}
    \displaystyle \min_{\mathbf{L}} \sum_{i=1}^n \left(S_{\mathbf{L}}^{\tilde{y}_i}\left(\mathbf{x}_i\right) - S_{\mathbf{L}}^{y_i}\left(\mathbf{x}_i\right)\right) + \lambda \norm{\mathbf{L}}_1
    \label{eq:CAMELnoPL}
\end{equation}
\noindent Note that \eqref{eq:CAMELnoPL} is a minimization problem, where \eqref{eq:CAMELnoReg} is a maximization problem.  We simply multiplied the objective in \eqref{eq:CAMELnoReg} by $-1$ to turn it into an equivalent minimization problem before adding the regularization term.  Here, the $l$-1 norm is taken element-wise on $\mathbf{L}$, that is $\norm{\mathbf{L}}_1 = \sum_{i=1}^{m'}\sum_{j=1}^{m}\left|\mathbf{L}^{i,j}\right|$.  Higher settings for the hyperparameter $\lambda$ force elements of $\mathbf{L}$ to exactly zero, preferring sparse solutions to more dense ones.  More sparse solutions may then be found that fit less to the observed data, thus reducing the risk of over-fitting.  This also has a more practical benefit.  Tens, hundreds, even thousands of features can be extracted from EHR data. It can be difficult to tell, a apriori, which are useful to model patient relationships.  If many features are used to represent patients ($m$ is large) and the learned model is dense, then a clinician has to consider many, potentially irrelevant, features to understand how the model is making decisions.  If $\mathbf{L}$ consists of a large number of zeros, then the learned metric only utilizes a few features, giving the clinician a concise model to interpret.  We call our gradient descent method to solve \eqref{eq:CAMELnoPL} CAMEL.

\subsection{Incorporating Confidence Labels}
CAMEL attempts to maximize the correct class confidence score for each training patient using only class labels.  As stated previously, we wish to also use confidence labels provided by clinicians to reduce the overall cost of obtaining expert supervision.  The most obvious way of incorporating confidence labels would be to ensure that the confidence score for an observed patient matches the confidence label the clinician provides.  However, in practice, confidence labels tend to contain a great deal of noise.  It has been shown that humans tend to find it difficult to accurately produce exact numerical assessments on objects and are much better suited to provide simpler forms of feedback such as class labels or relative comparisons \cite{kendall1990rank, stewart2005absolute}.  Because of this, bolstering CAMEL with the exact values of the confidence labels can introduce unwanted noise.  

Instead, we choose to simplify the labels by assuming that, while the exact value of a confidence label is noisy, its value compared to others of the same class is not (or at least reasonably less noisy).  For instance, if a clinician gives labels $c_a = 0.65$ and $c_b = 0.95$ we only take that to mean that the labeler is more confident in $y_b$ than $y_a$ with no consideration into by how much.  By this assumption, we create a ranking $\mathcal{R}_{\mathcal{C}}$ of patients such that $\left(\mathbf{x}_a,\mathbf{x}_b\right) \in \mathcal{R}_{\mathcal{C}}$ if and only if $c_a > c_b$ and $y_a = y_b$.  From $\mathcal{R}_{\mathcal{C}}$ we can induce a set of constraints to be imposed on \eqref{eq:CAMELnoPL} that allows us to incorporate the information contained in the confidence labels:
\begin{equation}
  \begin{array}{rll}
    \displaystyle \min_{\mathbf{L}} & \multicolumn{2}{l}{\displaystyle \sum_{i=1}^n \left(S_{\mathbf{L}}^{\tilde{y}_i}\left(\mathbf{x}_i\right) - S_{\mathbf{L}}^{y_i}\left(\mathbf{x}_i\right)\right) + \lambda \norm{\mathbf{L}}_1} \\ 
    \mathrm{s.t} & \displaystyle \forall_{\left(\mathbf{x}_a,\mathbf{x}_b\right) \in \mathcal{R}_{\mathcal{C}}} & \left(S_{\mathbf{L}}^{y_a}\left(\mathbf{x}_a\right) - S_{\mathbf{L}}^{\tilde{y}_a}\left(\mathbf{x}_a\right)\right) > \\
& & \left(S_{\mathbf{L}}^{y_b}\left(\mathbf{x}_b\right) - S_{\mathbf{L}}^{\tilde{y}_b}\left(\mathbf{x}_b\right)\right) 
  \end{array}
    \label{eq:CAMEL-CL-Const}
\end{equation}
\noindent The added constraints ensure that the approximated confidence score for $\mathbf{x}_a$ is greater than $\mathbf{x}_b$ for all $\left(\mathbf{x}_a,\mathbf{x}_b\right) \in \mathcal{R}_{\mathcal{C}}$.  In other words, it tries to ensure the confidence scores adhere to $\mathcal{R}_{\mathcal{C}}$.  In practice, it is unlikely that all of the constraints can be satisfied, so we opt to solve a similar, unconstrained optimization:
\begin{equation}
  \begin{array}{rll}
    \displaystyle \min_{\mathbf{L}} & \multicolumn{2}{l}{\displaystyle \sum_{i=1}^n \left(S_{\mathbf{L}}^{\tilde{y}_i}\left(\mathbf{x}_i\right) - S_{\mathbf{L}}^{y_i}\left(\mathbf{x}_i\right)\right) + \lambda_1\norm{\mathbf{L}}_1} \\ 
    & + \lambda_2\hspace{-1.35em}\displaystyle \sum_{\left(\mathbf{x}_a,\mathbf{x}_b\right) \in \mathcal{R}_{\mathcal{C}}} & \hspace{-2.1em}[S_{\mathbf{L}}^{\tilde{y}_a}\left(\mathbf{x}_a\right) - S_{\mathbf{L}}^{y_a}\left(\mathbf{x}_a\right) -\\
& & \hspace{-1.75em}S_{\mathbf{L}}^{\tilde{y}_b}\left(\mathbf{x}_b\right) + S_{\mathbf{L}}^{y_b}\left(\mathbf{x}_b\right)]_+
  \end{array}
    \label{eq:CAMEL-CL-Unconst}
\end{equation}
\noindent Here $[\cdot]_+$ is the hinge-loss function (equivalent to $\max\left(0,\cdot\right)$). The last term in \eqref{eq:CAMEL-CL-Unconst} is zero when the corresponding constraint in \eqref{eq:CAMEL-CL-Const} is satisfied and positive when it is not.  In short, if an observed object should have a higher confidence score than others because the labeler is more confident in its class label, that object's contribution to the objective is increased by a factor of $\lambda_2$ for each observed object its confidence score should be higher than.  If an observed object's confidence score is too high, its contribution is similarly decreased by a factor of $\lambda_2$.

By introducing the ranking term into the objective, we also introduce an additional hyperparameter $\lambda_2$.  Much like higher values of $\lambda_1$ increase the influence of regularization in the objective, higher values of $\lambda_2$ put a heavier emphasis on ordering the confidence scores according to $\mathcal{R}_{\mathcal{C}}$.  As such, care must be taken to properly set the hyperparameters to balance fit to the class labels, fit to the confidence labels, and sparsity.  We call our gradient descent method for solving \eqref{eq:CAMEL-CL-Unconst} CAMEL-CL.


\section{Related Work}
\label{sec:relWork}
Over the previous two decades, numerous metric learning methods have been developed, the two most similar to CAMEL being Large-Margin Nearest Neighbors (LMNN) \cite{weinberger2009distance}, and Metric Learning for Kernel Regression (MLKR) \cite{weinberger2007metric}. LMNN learns a metric from class labels to be used in a nearest neighbor classifier.  To do so, the authors formulate an optimization that ``pulls'' objects close to each other that are of the same class, and ``pushes'' objects of different classes away to ensure that the $k$ nearest neighbors to all observed instances are of the same class.  CAMEL uses similar notions of push and pull energies, but does so over all points, not just $k$ neighbors.  Also, LMNN directly optimizes their metric using notions of distance between objects, while CAMEL leverages a Gaussian kernel as a similarity measure that naturally induces confidence scores.  Learning such a similarity measure is similar to what is done for MLKR.  However, MLKR, as the name implies, learns a kernel to be used for regression.  If MLKR is applied to the problem setting considered in this work, it would fit directly to the confidence labels without consideration of the inherent noise within.  In Sec. \ref{sec:experiments} we compare CAMEL and CAMEL-CL to both LMNN and MLKR in our experimental evaluation.

Two previous works have considered auxiliary labels similar to the confidence labeled considered in our problem setting.  Both use \emph{probabilistic labels} that indicate how likely an object belongs to a class. The first work \cite{nguyen2011learning, nguyen2014learning} introduced the problem of learning from probabilistic auxiliary labels and formulates a method that uses the popular Support Vector Machine (SVM) framework.  Their method learns linear classifier by solving an optimization problem that balances two energies: standard SVM classification hinge loss and a function that encourages the model to rank the objects by their probabilistic labels.  By including the ranking energy, their method can produce an SVM in which objects with high probabilistic labels are farther from the decision hyperplane than those with low probabilistic labels.  While our method uses a similar ranking energy, it is based on an intuitive, explicitly defined confidence score. In addition, their learned model is a linear transformation into a single dimension, and our model is a more expressive multidimensional metric.  Finally, our method induces sparsity in order to improve accuracy and enhance interpretability, while their method learns a dense model.  Both \cite{nguyen2011learning} and \cite{nguyen2014learning} contain experimental evaluation on clinical data sets.  However, they provide no analysis into what can be interpreted from their model, opting only to show results pertaining to accuracy of inference.  Another work uses Gaussian Process Regression \cite{Peng2014Learning} to learn a classifier from the probabilistic labels, but because the SVM-based method is most similar to our work in both methodology and application, we compare CAMEL and CAMEL-CL to SVM-Combo from \cite{nguyen2014learning} in Sec. \ref{sec:experiments}.

\section{Experimental Evaluation}
\label{sec:experiments}
\begin{table}[t]
  \begin{centering}
  \begin{tabular}{|l|l|l|l|l|}
    \hline
    \textbf{Expert} & \textbf{\# Pos} & \textbf{\# Neg} & \textbf{Mean CL Pos (STD)} & \textbf{Mean CL Neg (STD)}\\
    \hline
    1 & 76 & 473 & 0.583 (0.151)& 0.294 (0.089)\\
    \hline
    2 & 143 & 428 & 0.656 (0.377)& 0.114 (0.202)\\
    \hline
  \end{tabular}
\caption{Summary Statistics for Expert-Labeled Data}
\label{tab:ExpertStats}
\end{centering}
\vspace{-2em}
\end{table}
\begin{table*}[t]
  \begin{centering}
  \begin{tabular}{|l|l|l|l|l|}
    \hline
    \textbf{Name} & \textbf{Supervision} & \textbf{Hyperparameters} & \textbf{Implementation Used} & \textbf{Brief Description}\\
    \hline
    \textbf{Orig} & None & None & Our own & Gaussian kernel (bandwidth = 1) applied to original feature space \\
    \hline
    \textbf{SVM} & Class Labels & Weight on hinge-loss & LIBLINEAR \cite{fan2008liblinear} & Standard linear SVM classification \\
    \hline
    \textbf{LMNN} & Class Labels & Number of nearest neighbors& From author's website &  Metric learning method for nearest neighbor classification\\
    \hline
    \textbf{CAMEL} & Class Labels & $\lambda$ & Our own & Our algorithm for solving \eqref{eq:CAMELnoPL} \\
    \hline
    \textbf{LASSO} \cite{tibshirani1996regression}& Confidence Labels & Weight on $l$-1 norm & MATLAB Stats \& ML & $l$-1 norm regularized linear regression \\
    \hline
    \textbf{MLKR} & Confidence Labels & Dimension of metric projection & From author's website & Metric learning for kernel regression \\
    \hline
    \textbf{SVM-Combo} & Class \& Confidence Labels & Weight on classification loss & Provided by author & Standard linear SVM classification with penalty that enforces \\
    &  & and weight ranking loss & & ranking of patients by confidence labels\\
    \hline
    \textbf{CAMEL-CL} & Class \& Confidence Labels & $\lambda_1$ and $\lambda_2$ & Our own & Our algorithm for solving \eqref{eq:CAMEL-CL-Unconst} \\
    \hline
  \end{tabular}
  \caption{Methods used in experimental evaluation}
  \label{tab:Methods}
  \end{centering}
\vspace{-2em}
\end{table*}
To evaluate CAMEL and CAMEL-CL, we performed experiments on real-world clinical data, the results of which we discuss in this section.  We begin by first describing the data set used.  Then, we outline how the experiments were performed.  Next, we present and discuss quantitative results comparing both CAMEL methods to related, current methods.  Finally, we qualitatively analyze the models learned by CAMEL by looking at how it uses patient data to make inferences.





\subsection{Data Set Description}
\label{sec:Data}
Our experiments were performed on data extracted from the Post-Surgical Cardiac Patient (PCP) Database in combination with supervision provided by clinicians indicating whether a patient is at risk for Heparin-Induced Thrombocytopenia (HIT).  A lengthy description of this data can be found in \cite{nguyen2014learning} and further specifics can be found in \cite{hauskrecht2010conditional, valko2010feature, hauskrecht2013outlier}.  Here, we provide a shorter summary of our view of the data.  From the PCP Database 4,486 unique EHRs were chosen.  From these over 51,000 patient-state instances were extracted using 24-hour segmentation.  Uniformly random sampling from the pool of patient instances would result in an overwhelmingly disproportionate number of negative labels for HIT.  Because there was a finite budget in obtaining supervision, a stratification procedure was used to bias sampling towards patients that could be at risk for HIT.  Using this procedure, patient instances were chosen to be labeled by three experts in clinical pharmacology.  

The experts were asked two questions for each patient instance: ``How strongly does the clinical evidence indicate that the patient is at risk of HIT?'' and ``Assume you have received an HIT alert for this patient. To what extent you agree/disagree with the alert?''.  For the first question, the experts were prompted for a number between 0 and 100, which we normalize to $[0,1]$ and use as confidence labels.  For the second question, the experts were prompted to for one of four ordinal categories ranging from ``strongly agree'' to ``strongly disagree'' from which we derive binary class labels between the ``agree'' and ``disagree'' categories.  Both our experiments and the results reported in \cite{nguyen2014learning} indicate that the confidence labels provided by the third expert are prohibitively noisy.  To more properly showcase the potential utility of confidence labeling, we omit lengthy discussion of results from experiments using the third expert's supervision, and note that all methods that utilize the third expert's confidence labels alone perform poorly.  Furthermore, the two methods that utilize both class labels and confidence labels gain no benefit from the confidence labels.  Statistics summarizing the two other experts' feedback used in our evaluations (number of positively labeled patients, mean confidence labels, etc.) can be found in Table \ref{tab:ExpertStats}.

From the EHRs of the selected patient instances, 50 features were extracted to form feature vectors characterizing each patient.  These features measure both trends and static measurements in one of five \emph{attributes}: Heparin administration record (features 1-4), hemoglobin count (5-18), white blood cell count (19-31), platelet count (32-45), and major heart surgeries (46-49). For example, platelet count features include: ``latest platelet value taken'', ``difference between last two platelet values taken'', and ``overall trend in platelet values''.
  
\subsection{Experimental Methodology}
\label{sec:Consider}
The methods used in this evaluation are listed in Table \ref{tab:Methods}.  These methods were chosen as a sample of current techniques that learn linear models or metrics from one or both forms of supervision considered in this work.  The ``Orig'' method uses no supervision and provides us with a rudimentary baseline in our evaluation.  All methods can produce confidence scores:  The classification models can be interpreted to have uncertainty measures in class predictions, and the real-valued predictions from the regression models can be taken as confidences.  Thus, we evaluate the accuracy of each method using the AUROC of predictions on a held-out test set. 

We performed separate but identical experiments on each experts' supervision.  From the pool of selected patient instances we randomly selected 100 patients to be the train set, and split the remaining patients randomly into evenly-sized test and validation sets.  This was done 20 times to form 20 trials.  For each trial, an increasing number of the 100 training points were used to train the models in the evaluation (10, 20,...100).  We did this to assess each method as a function of the amount of obtained supervision.  For each training partition, hyperparameter settings for each method were chosen to be those that maximized AUROC on the validation set.

\subsection{Discussion}
\label{sec:Discussion}
\begin{figure*}[t]
\hspace{-0.75em}
  \centering
\begin{subfigure}{0.5\textwidth}
    \centering
    \includegraphics[width=1.0\textwidth]{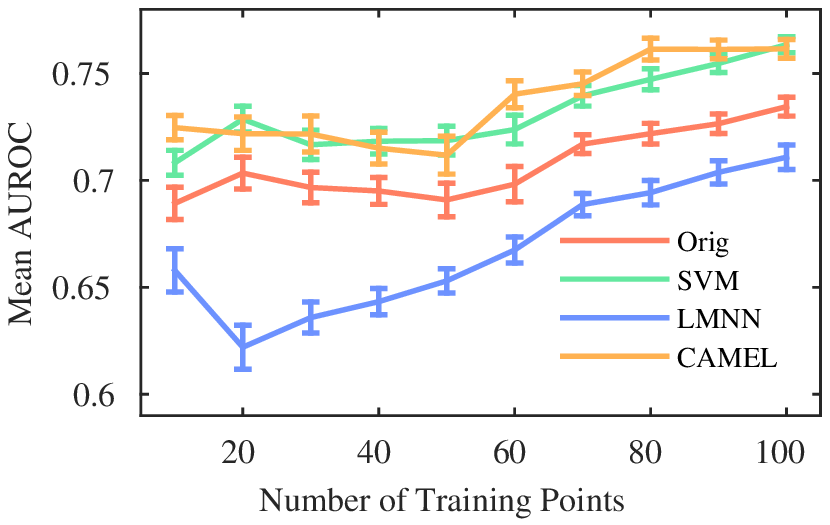}
  \end{subfigure}%
  \hfill %
  \begin{subfigure}{0.5\textwidth}
    \centering
    \includegraphics[width=1.0\textwidth]{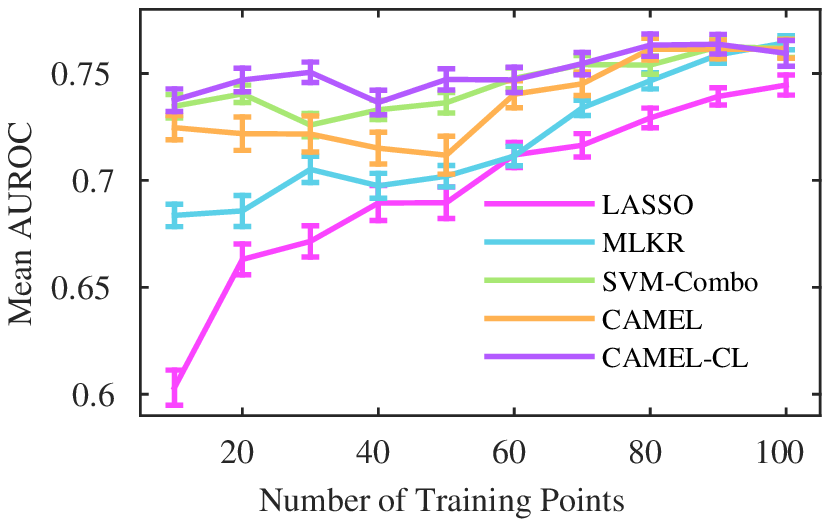}
  \end{subfigure}%

\hspace{-0.75em}
\begin{subfigure}{0.5\textwidth}
    \centering
    \includegraphics[width=1.0\textwidth]{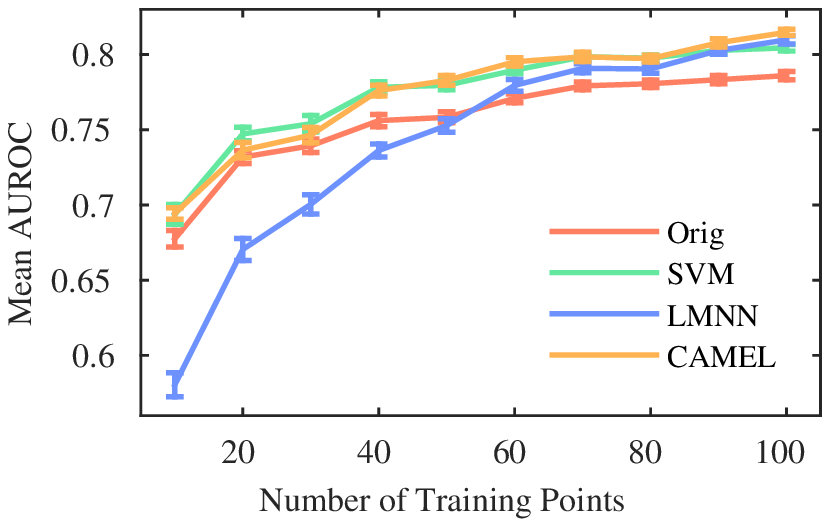}
  \end{subfigure}%
  \hfill %
  \begin{subfigure}{0.5\textwidth}
    \centering
    \includegraphics[width=1.0\textwidth]{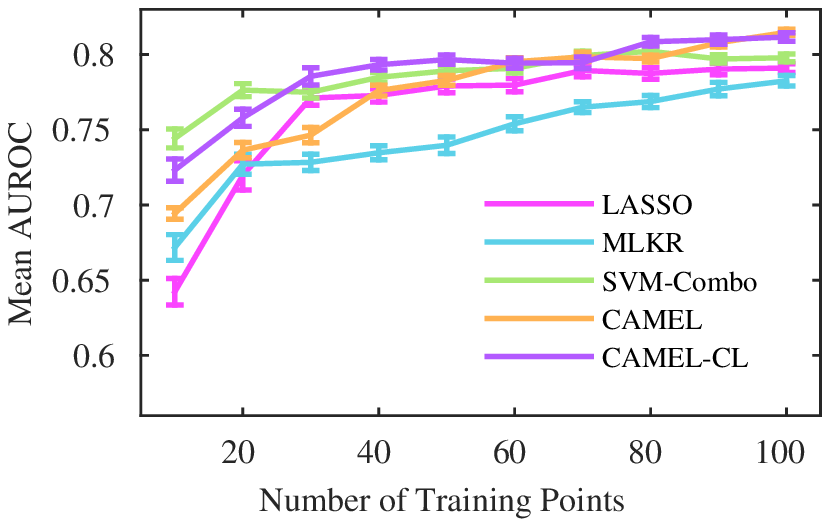}
  \end{subfigure}%
    \caption{Number of Training Points vs. AUROC on Test Set (Row 1 = Expert 1, Row 2 = Expert 2)}
    \label{fig:AUCs}
    \vspace{-1em}
\end{figure*}
Figure \ref{fig:AUCs} shows plots of the AUROC values on the test set as function of the number of training points for all methods used in our experiments.  The error bars represent a 95\% confidence interval. The top two plots are for Expert 1's supervision, and the bottom two are for Expert 2's.  The left plots show results for methods that do not use confidence labels and the right plots are methods that do (with CAMEL in both for comparison between the plots).  Overall, the results were similar for both clinicians. All methods achieved better results using Expert 1's labels when given few training points.  Although, as more training points were added, all methods were able to improve more using Expert 2's supervision.


In terms of classification accuracy, CAMEL and CAMEL-CL performed as well or better than many of the competing methods, especially as more training instances were used.  LMNN and MLKR learn metrics like CAMEL, but do not include regularization in their optimizations.  Because of this, they are prone to over-fit to train sets, resulting in poor generalization, especially when there are few training instances.  LASSO includes sparsity-inducing regularization, but learns a simple single-dimensional, linear model.  Also, like MLKR, LASSO fits directly to the confidence labels.  Because the exact values of the confidence labels contain a great deal of noise, LASSO and MLKR are unable to learn models that produce accurate confidence values on the test instances.

The most competitive models to CAMEL and CAMEL-CL, using the same supervision, were SVM and SVM-Combo, respectively.  The SVM methods learn dense, single-dimensional models, so they provide a meaningful basis of comparison to our methods.  CAMEL performed at least as well and sometimes significantly better than SVM.  The same is true for CAMEL-CL and SVM-Combo, though SVM-Combo outperformed CAMEL-CL for 10 and 20 training instances using Expert 2's supervision. With only this exception, CAMEL and CAMEL-CL were able to achieve AUROCs as high or higher than the SVM methods, given the same supervision.  These results indicate that to our sparse, multidimensional models are able to more accurately predict whether a patient was at risk for HIT, than methods that learn dense and/or single-dimensional models.

This evaluation not only allowed us to compare our CAMEL methods to competing methods, but also enabled us to see the effect the confidence labels had on CAMEL.  The inclusion of confidence labels only improved the AUROC of CAMEL for both experts, never hindered it.  Using Expert 1's supervision, CAMEL-CL was able to achieve an AUROC with 10 training instances that could not be matched by CAMEL until it received 60.  Furthermore, the AUROC of the CAMEL-CL model trained on 10 instances was statistically as high as any model trained on any number of instances with a 95\% confidence.  This indicates that to learn an accurate predictive model, CAMEL-CL requires substantially fewer labeled instances.
\begin{table}[t]
  \begin{centering}
  \begin{tabular}{|l|l|l|l|l|}
    \hline
    & \multicolumn{2}{c|}{\textbf{Expert 1}} & \multicolumn{2}{c|}{\textbf{Expert 2}} \\ \hline
    & \multicolumn{1}{c|}{Sparsity} & \multicolumn{1}{c|}{Rank} & \multicolumn{1}{c|}{Sparsity} & \multicolumn{1}{c|}{Rank} \\ \hline 
    \hspace{-0.2em}\textbf{CAMEL}\hspace{-0.2em}& \hspace{-0.2em}0.794(0.241)\hspace{-0.2em}& \hspace{-0.2em}34.195(18.884)\hspace{-0.2em} & \hspace{-0.2em}0.754(0.237)\hspace{-0.2em} & \hspace{-0.2em}38.734(16.311)\hspace{-0.2em} \\ \hline
    \hspace{-0.2em}\textbf{CAMEL-CL}\hspace{-0.2em}& \hspace{-0.2em}0.896(0.286)\hspace{-0.2em}& \hspace{-0.2em}20.080(18.921)\hspace{-0.2em} & \hspace{-0.2em}0.883(0.203)\hspace{-0.2em} & \hspace{-0.2em}21.755(18.862)\hspace{-0.2em}\\
    \hline
  \end{tabular}
  \caption{Mean (STD) sparsity statistics over all experiments}
  \label{tab:SparStats}
  \end{centering}
  \vspace{-2em}
\end{table}

While prediction accuracy is important for a patient model, for it to be useful in a CDSS it must also be interpretable. In the remainder of this section, we assess the metrics produced by CAMEL and CAMEL-CL in terms of human-interpretability.  In our experiments $\mathbf{L}$ is a 49 by 49 matrix that transforms the patient instances from their original space to a multidimensional metric space by linearly combining their features.  Each row is a transformation to a different dimension in the metric space, and each column represents an individual feature's contribution to the model.  In essence, each element of $\mathbf{L}$ corresponds to one of 49 weights on one of the features (e.g. the 49 values in column 1 weigh \emph{Heparin on} feature).  If not sufficiently sparse, $\mathbf{L}$ could be difficult to interpret, as the model would use many of the features, multiple times, in numerous different combinations.  A sparse $\mathbf{L}$ that produces accurate inferences would use only a subset of the features in few, useful combinations, which could be easier to interpret than many, complex combinations. 

Table \ref{tab:SparStats} shows the mean ``sparsity'' of the metric parameter $\mathbf{L}$ produced by CAMEL and CAMEL-CL for all experiments.  We define our sparsity statistic as the number of zero-valued elements of $\mathbf{L}$ divided by the total number of elements.  A higher value means that the fewer features are being used fewer times in the model.  We can see that models learned by CAMEL contain a very large number of zero-valued elements, but CAMEL-CL is able to be even more selective in choosing features by leveraging the confidence labels. Also in Tab. \ref{tab:SparStats}, we include the mean row rank of the $\mathbf{L}$ matrices.  The row rank of a matrix is the number of linearly independent rows.  In our models, a lower rank indicates there is a more simple, lower-dimensional space that describes how the CAMEL models are making inferences.  The table shows that our methods are able to project the 49 dimensional patient instances into a lower-dimensional metric space in which accurate inferences can be made.  The low-rank property of $\mathbf{L}$ can be attributed to the fact that strict $l$-1 norm regularization often made many of the rows contain all zeros.  For some trials, our methods produced an $\mathbf{L}$ with as many as 45 rows that contained solely zeros. 
\begin{figure*}[t!]
  \centering
  \hspace{-2em}
    \includegraphics[width=1.0\textwidth]{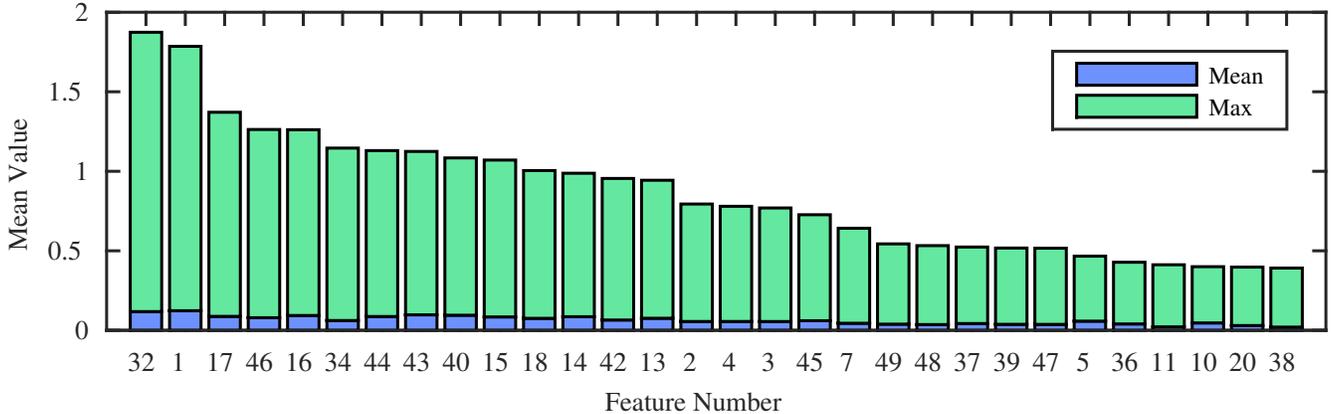}
    \vspace{-1.5em}
    \caption{Feature weight statistics (CAMEL-CL, Clinician 2)}
    \label{fig:FeatStats}
    \vspace{-1em}
\end{figure*}

While sparsity indicates the models are simple, it does not reveal how the features are being used.  More specifically, for a model to be interpretable, a clinician should be able to tell which features are being used, how much, and whether they are important on their own, or in tandem with others.  Figure \ref{fig:FeatStats} shows the mean and maximum absolute weight put on the top 30 features in the models produced by CAMEL-CL, averaged over all experiments done using Expert 2's supervision.  
In short, Fig. \ref{fig:FeatStats} displays the relative importance CAMEL-CL put on each feature. The clear top two features chosen by CAMEL-CL were ``last platelet value taken'' (32) and ``Heparin on'' (1). Clearly, whether a patient was given Heparin should influence whether they are at risk for HIT.  Thrombocytopenia is indeed the deficiency of platelets in blood \cite{warkentin2000impact}, thus the most recent value of platelet count intuitively should indicate risk of HIT.  Other top features include ``difference between the last and first hemoglobin level taken'' (17), and ``time since last major heart procedure'' (46). A downward trend in hemoglobin level could indicate bleeding, leading to low platelet counts, making feature 17 a potential indicator of HIT. The time from last heart procedure could also be important as it indirectly measures how long the patient was on heparin.  Note that the top four features all come from different attributes/lab values.  This indicates that CAMEL-CL chooses which feature in a group is most informative and emphasizes it the most, as to not include redundant information.  Also note that no feature measuring white blood cell count was featured prominently in the model.  This model choice is supported by the convention that white blood cell count is not commonly-used to indicate HIT.

\begin{figure*}[t!]
  \centering
    \includegraphics[width=1.00\textwidth]{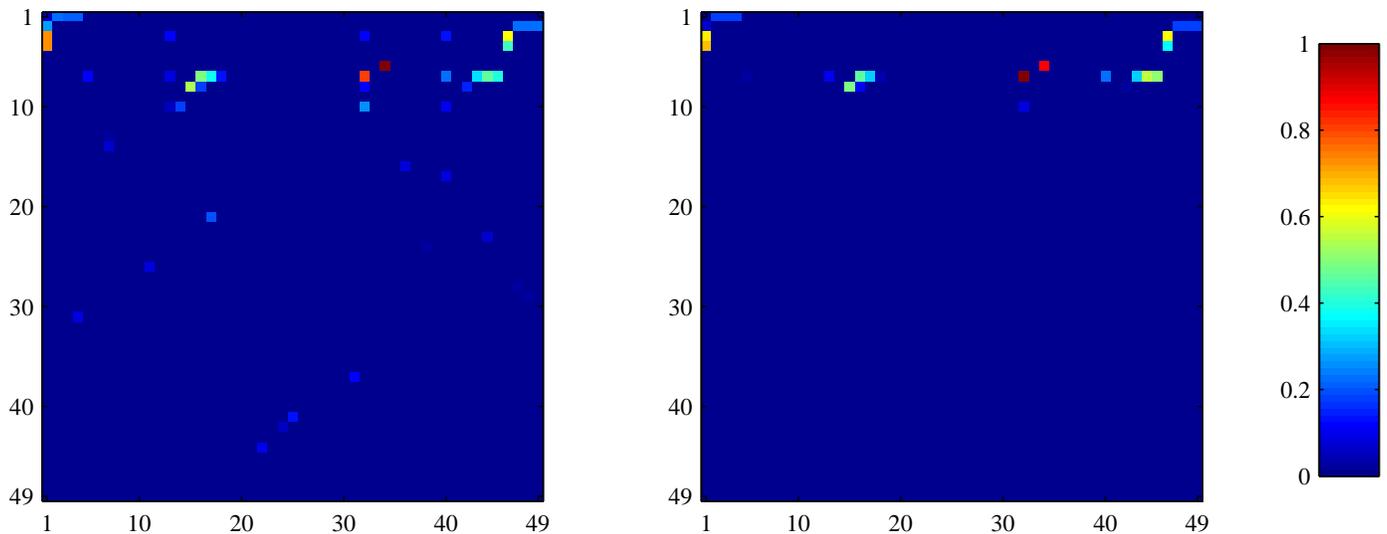}
    \vspace{-0.5em}
    \caption{Normalized absolute weights of CAMEL (left) and CAMEL-CL (right) $\mathbf{L}$ parameter}
    \label{fig:HeatMaps}
    \vspace{-1em}
\end{figure*}

Figure \ref{fig:HeatMaps} displays two heat maps using the normalized absolute values of $\mathbf{L}$.  Deep blue indicates a zero value, while deep red indicates the highest absolute value.  The two maps depict $\mathbf{L}$ for CAMEL (left) and CAMEL-CL (right) in one trial using the same 100 training instances.  Each row is a projection (weighing of the features) into a single dimension in the metric space. Thus, each row defines a different ``factor'' in which the metrics compare patients. The left heat map shows that the model produced by CAMEL is very sparse; it has mostly zero-valued elements (deep blue), a small number of small-valued elements (light blue), and an even smaller number of larger-valued elements (green, yellow, and red).  In total, this matrix has 25 rows composed entirely of zeros (i.e. the induced metric space is $\mathbf{L}$ is 24 dimensional).  However, many of the non-zero rows contribute very little to the overall model, as they contain only few, low-valued weights.  Most likely, these rows simply add noise, and detract from the interpretability of the model by unnecessarily increasing the complexity.  The heat map displaying the metric learned by CAMEL-CL, on the other hand, has many more zero-valued elements.  Most of the small-valued elements in the CAMEL model were pushed to exactly zero in the CAMEL-CL model.  In fact, every element of $\mathbf{L}$ learned by CAMEL-CL after the tenth row contains a zero, resulting a simpler, rank ten matrix.  Because CAMEL-CL metric was given confidence labels in addition to the class labels, it was able to more accurately determine the few feature combinations that modeled patients well.  

\section{Conclusion and Future Work}
\label{sec:conclusion}
In this work, we developed a method called CAMEL that produces sparse, multidimensional, classification models that can perform inference on patient Electronic Health Records (EHRs) for use in Clinical Decision Support Systems (CDSSs).  For an inference model to be used in a CDSS, it must be both accurate and able to be interpreted by a clinician.  CAMEL was designed specifically with these qualities in mind.  In order to combat the necessarily high cost of obtaining expert clinical supervision needed to learn an accurate model, we formulated a version CAMEL that can incorporate auxiliary confidence labels.  In our experiments, we showed that CAMEL can produce models at least as accurate as others we tested, and CAMEL bolstered with confidence labels can produce models as accurate as any tested with using as few as 10\% of the training instances as the other models.  The qualitative analysis that followed highlighted the fact that CAMEL produces models that include few important ``factors'' composed of small subsets of the EHR features.  Because CAMEL induces sparsity, it is able produce simple, concise patient models, potentially enabling clinicians to more clearly interpret how it makes decisions.

There are multiple avenues of future work we will explore.  First, CAMEL-CL uses simple pair-wise constraints to enforce a ranking of patients.  Current methods in learning to rank \cite{liu2009learning} have evolved to use more intricate and effective means to ensure a ranking of objects.  We will investigate some of the ideas from these methods for inclusion into CAMEL.  Second, the features we extracted from the EHRs contained numerous binary features that were deemed important by CAMEL and CAMEL-CL.  This indicates that our methods determined that dividing the patients into certain subpopulations produced better results.  It is likely that patients within these subpopulations should be modeled in different ways.  We will investigate learning different metrics defined on separate subspaces of the data to model the different dynamics of each subpopulation.  Finally, we performed our experiments using each expert's supervision separately.  We will investigate pooling supervision from all experts to learn a combined model, potentially allowing it to represent a more universal view of how clinicians view patients.

\section*{Acknowledgment}
\label{sec:acknowledgement}
The collection of the data set used in this work was supported by National Institutes of Health (NIH) grants R01-LM010019 and R01-GM088224.

\bibliographystyle{./IEEEtran}
\bibliography{./IEEEabrv,./CASTLE}
\end{document}